\documentclass[conference]{IEEEtran}
\usepackage{subcaption}
\IEEEoverridecommandlockouts
\usepackage{cite}
\usepackage{amsmath,amssymb,amsfonts}

\usepackage{graphicx}
\usepackage{textcomp}
\usepackage{xcolor}
\usepackage{algorithm}
\usepackage{algpseudocode}
\usepackage{algorithmicx}
\usepackage{booktabs}
\usepackage{caption}

\usepackage{nameref}

\usepackage{multirow}
\def\BibTeX{{\rm B\kern-.05em{\sc i\kern-.025em b}\kern-.08em
    T\kern-.1667em\lower.7ex\hbox{E}\kern-.125emX}}
\begin{document}
\bibliographystyle{IEEEtran}
\title{CFPFormer: Cross Feature-Pyramid Transformer Decoder for Medical Image Segmentation\\
\thanks{* These authors contributed equally to this work.}
}
\author{
\IEEEauthorblockN{1\textsuperscript{st} Hongyi Cai*}
\IEEEauthorblockA{Faculty of Computer Science\\
Universiti Malaya\\
Kuala Lumpur, Malaysia\\
xcloudfance@gmail.com}
\and
\IEEEauthorblockN{2\textsuperscript{nd} Mohammad Mahdinur Rahman*}
\IEEEauthorblockA{Faculty of Computer Science\\
Universiti Malaya\\
Kuala Lumpur, Malaysia\\
rahmanmahdinur@gmail.com}
\and
\IEEEauthorblockN{3\textsuperscript{rd} Wenzhen Dong}
\IEEEauthorblockA{Department of Mechanical and\\
Automation Engineering\\
The Chinese University of Hong Kong\\
Hong Kong, China\\
dongwz@link.cuhk.edu.hk}
\and
\IEEEauthorblockN{4\textsuperscript{th} Jingyu Wu}
\IEEEauthorblockA{Fuzhou University of\\
International Studies and Trade\\
Fujian, China\\
2303669172@qq.com}
}

\maketitle

\begin{abstract}
  Feature pyramids have been widely adopted in convolutional neural networks and transformers for tasks in medical image segmentation. However, existing models generally focus on the Encoder-side Transformer for feature extraction. We further explore the potential in improving the feature decoder with a well-designed architecture. We propose Cross Feature Pyramid Transformer decoder (CFPFormer), a novel decoder block that integrates feature pyramids and transformers. Even though transformer-like architecture impress with outstanding performance in segmentation, the concerns to reduce the redundancy and training costs still exist. Specifically, by leveraging patch embedding, cross-layer feature concatenation mechanisms, CFPFormer enhances feature extraction capabilities while complexity issue is mitigated by our Gaussian Attention. Benefiting from Transformer structure and U-shaped connections, our work is capable of capturing long-range dependencies and effectively up-sample feature maps. Experimental results are provided to evaluate CFPFormer on medical image segmentation datasets, demonstrating the efficacy and effectiveness. With a ResNet50 backbone, our method achieves 92.02\% Dice Score, highlighting the efficacy of our methods. Notably, our VGG-based model outperformed baselines with more complex ViT and Swin Transformer backbone.
\end{abstract}


\section{Introduction}
The field of medical image analysis has been revolutionized by deep learning techniques, particularly convolutional neural networks (CNNs) like U-Net \cite{ronneberger_u-net_2015}. CNNs excel at capturing local features and spatial hierarchies, making them highly effective for tasks such as image segmentation. However, as networks deepen, they might struggle to maintain global context, which is crucial for accurate diagnosis, treatment planning, and disease monitoring in healthcare applications.

Transformer-based methods, on the other hand, have demonstrated remarkable capabilities in capturing long-range dependencies and global context in image processing \cite{liu2021swin, carion_end--end_2020}. In medical image segmentation, majority of precedent works \cite{dosovitskiy_image_2021, tragakis_fully_2023, cao_swin-unet_2023} have enlightened transformer encoders, processing 2D image patches with positional encodings as input sequences, allowing them to model relationships across the entire image effectively. While much research has focused on improving encoder architectures to combine the strengths of CNNs and transformers, less attention has been paid to the decoder component. Decoders play a crucial role in expanding on the features extracted by encoders, particularly in tasks requiring detailed output such as medical image segmentation. Current U-Net variants predominantly employ transpose convolutions or bilinear upsampling in decoders \cite{oktay2018attention, schlemper_attention_2019}. However, these operations exhibit two critical limitations:
\begin{enumerate}
    \item Shallow spatial features from early encoder layers are merely concatenated with deep semantic features via skip connections, leading to suboptimal multi-scale interaction;
    
    \item Localized receptive fields of convolutions fail to model long-range anatomical dependencies (e.g., curvilinear vessels or organ boundaries), while global self-attention incurs prohibitive $\mathcal{O}(N^2)$ complexity for high-resolution medical images.
\end{enumerate} .

Motivated by these challenges and recognizing the potential for improvement in decoder architectures, we introduce a novel Cross Feature Pyramid Transformer (CFP) block acting as the decoder of segmentation models. By incorporating patch embedding and a modified attention mechanism, the CFP block aims to enhance the model's capacity to capture both fine details and global context, which are essential for medical image analysis tasks.

CFPFormer is hierarchical and incorporates feature pyramids, allowing fine-grained features from various layers to pass to the decoder. This multi-scale approach enables the model to capture features at different levels of abstraction. However, the effective utilization of this multi-scale knowledge requires proper encoding and prioritization. To address this, we introduce two key components: Feature Re-encoding (FRE) and Gaussian Attention.The CFP architecture is designed to be flexible, integrating seamlessly with various network backbones as encoders. This versatility makes our approach well-suited for diverse medical imaging applications, as will be demonstrated in subsequent sections.

Feature Re-encoding reassembles outputs from image encoders and adjusts them to fit into decoder layers, unfolding the latent potential of decoder-based models. FRE ensures that the output from different scales is properly integrated and represented in a format suitable for the decoder's processing.

A key innovation in our architecture is the utilization of Gaussian Attention mechanisms within the CFP block.  We propose Axially-Decomposed Gaussian Attention (GA), which reformulates global self-attention into row-wise and column-wise computations with a distance-decaying prior. This design aligns with the directional continuity of anatomical structures (e.g., elongated cardiac walls), reducing complexity from $O(H^2W^2)$ to $O(HW(H+W))$ while suppressing noise from irrelevant regions. Furthermore, this attention mechanism is precomputed, making it more computationally efficient than traditional self-attention approaches. This attention mechanism is designed to decay attention along a curve, capturing long-range dependencies without the high computational cost of pixel-level attentions.

The specific contributions of the proposed CFPFormer decoder block include:

\begin{itemize}
    \item We propose the CFPFormer mechanism, a hierarchical feature pyramid approach that effectively decodes long-range details from encoders by leveraging concepts from Vision Transformers, allowing for efficient multi-scale feature integration and processing.
    \item A modified Gaussian attention mechanism applied over rows and columns, enhancing the decoded feature map and contributing to improved performance.
    \item The introduction of Feature Re-encoding (FRE), which reassembles outputs from image encoders and adjusts them to fit into decoder layers, unfolding the latent potential of decoder-based models.
    
\end{itemize}

\section{Related Works}

\subsection{Convolutional Neural Networks (CNNs)}
Convolutional Neural Networks (CNNs) have been the cornerstone of medical image analysis for the past decade. The U-Net architecture \cite{ronneberger_u-net_2015}, with its encoder-decoder structure and skip connections, has set benchmarks in various segmentation tasks. Subsequent improvements like U-Net++ \cite{zhou2018unet} and UNet 3+ \cite{huang2020unet3fullscaleconnected} have further enhanced segmentation accuracy through dense skip connections and full-scale deep supervision.

CNNs excel at capturing local features and spatial hierarchies through their convolutional layers and pooling operations \cite{app132111657}. The hierarchical nature of CNNs allows them to learn features at multiple scales, crucial for handling the diverse sizes of anatomical structures in medical images \cite{Litjens_2017}. Recent advancements in CNN architectures for medical imaging include the attention-guided DenseUNet \cite{https://doi.org/10.1049/iet-ipr.2019.1527} and the Residual UNet \cite{Zhang_2018}, which incorporate residual learning to facilitate training of deeper networks.

However, CNNs face limitations in capturing long-range dependencies due to their localized receptive fields. Techniques like dilated convolutions \cite{yu2016multiscalecontextaggregationdilated} and atrous spatial pyramid pooling \cite{atlc} have been proposed to expand the receptive field without increasing computational cost, but these still struggle with truly global context modeling.

\subsection{Transformers in Vision}
Transformers, originally designed for natural language processing \cite{vaswani_attention_2023}, have recently made significant inroads in computer vision. The Vision Transformer (ViT) \cite{dosovitskiy_image_2021} demonstrated the potential of pure transformer architectures in image classification by treating image patches as tokens. This approach enables modeling of long-range dependencies, addressing a key limitation of traditional CNNs.

The success of ViT has spawned numerous variants optimized for vision tasks. DeiT \cite{touvron2021trainingdataefficientimagetransformers} introduced a data-efficient training strategy for vision transformers, while Swin Transformer \cite{liu2021swin} proposed a hierarchical structure with shifted windows, making it more suitable for dense prediction tasks like segmentation. In the medical domain, TransUNet \cite{chen_transunet_2021} and UNETR \cite{hatamizadeh2021unetrtransformers3dmedical} have shown promising results by combining transformer modules with CNN-like architectures.

However, the lack of inductive biases for spatial data and high computational requirements pose challenges in medical image analysis \cite{cai2024efficientvitmultiscalelinearattention}. These limitations have motivated research into more efficient transformer architectures and hybrid approaches.

\subsection{Limitations and Hybrid Approaches}
While CNNs struggle with long-range dependencies, transformers face challenges in capturing fine-grained local features efficiently\cite{wu2021cvtintroducingconvolutionsvision, nonlocal, liu2021swin}. To address these limitations, hybrid architectures combining CNNs and transformers have emerged. 

Models like TransUNet \cite{chen_transunet_2021} use a CNN encoder for feature extraction followed by a transformer module for global context modeling. SwinUNet \cite{cao_swin-unet_2023} adapts the Swin Transformer for medical image segmentation, incorporating a U-Net-like structure. CoTr \cite{xie_cotr_2021} proposes a hybrid CNN-Transformer model for volumetric medical image segmentation, leveraging the strengths of both architectures.

The nnFormer \cite{zhou2022nnformerinterleavedtransformervolumetric} introduces a pure transformer model for volumetric medical image segmentation, demonstrating competitive performance with carefully designed inductive biases. Another interesting approach is the Pyramid Vision Transformer (PVT) \cite{wang2021pyramidvisiontransformerversatile}, which creates a hierarchical structure similar to CNNs but uses transformer blocks at each scale.

These hybrid approaches aim to balance the strengths of both architectures, crucial for tasks requiring both detailed spatial information and global context understanding.



\subsection{Efficiency in Vision Transformers}
The high computational cost of vision transformers has spurred research into more efficient architectures. EfficientFormer \cite{li2022efficientformervisiontransformersmobilenet} introduces a dimension-consistent design and progressive patch embedding to reduce computational complexity. Other approaches to improve efficiency include Mobile-Former \cite{chen2022mobileformerbridgingmobilenettransformer}, which combines the efficiency of MobileNets with the expressive power of transformers, and Efficient Vision Transformer \cite{graham2021levitvisiontransformerconvnets}, which uses a multi-stage design and token mixing to reduce computation. PatchFormer \cite{cheng2022patchformerefficientpointtransformer} introduces an efficient patch-wise attention mechanism. These efficient designs are particularly relevant for medical image analysis, where computational resources may be limited and real-time processing is often desired.





\section{Method}
\textbf{Decoder for Downstream Tasks} Our model architecture is established upon the structure of Transformer-Decoder. Therefore, The architecture of CFPFormer aims to enhance the integration of information between Encoder and Decoder layers, which may result in more comprehensive use of encoded features.. At the core of this architecture lies the Cross Feature Pyramid (CFP) Block, which incorporates three key innovations: Gaussian Attention, Feature Re-encoding and Cross-Layer Feature Integration. These components are designed to enhance the model's ability to capture complex spatial relationships, integrate information across different scales, and mitigate the loss of dense information during downscaling.

\begin{figure*}[ht!]
    \centering
    \includegraphics[width=1.0\linewidth]{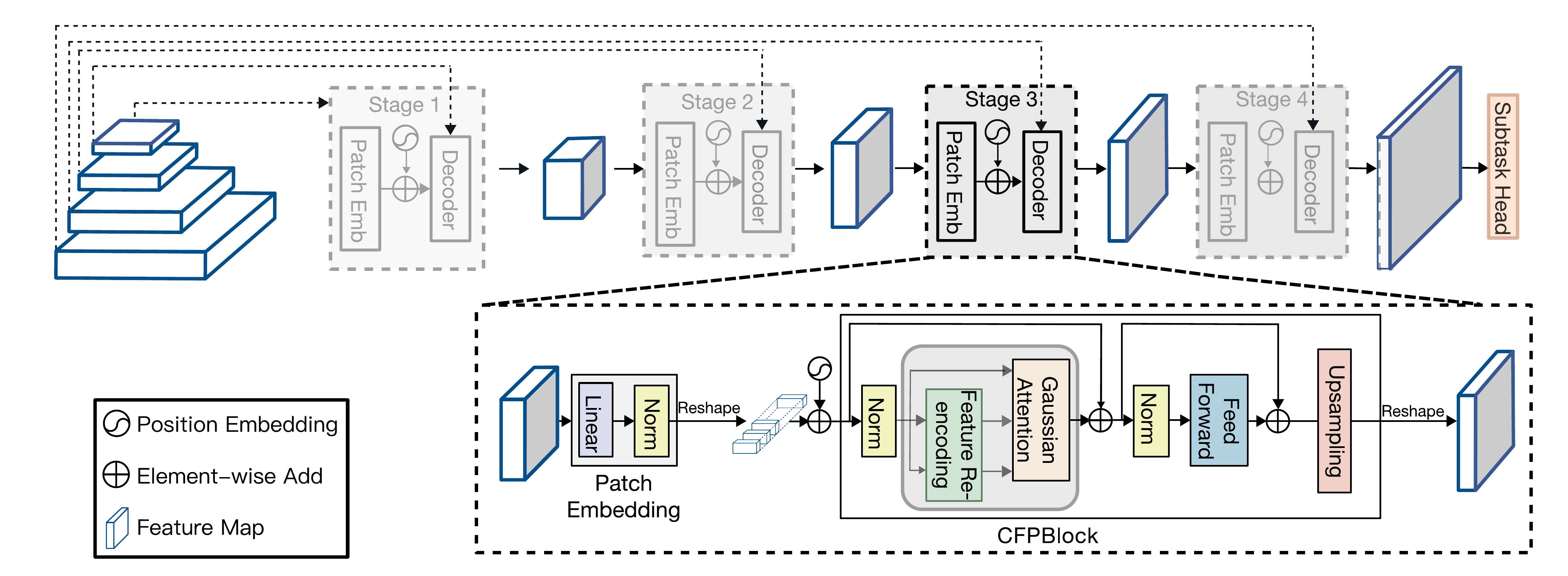}
    \caption{Overall architecture of Cross Feature Pyramid transformer decoder blocks: The input image undergoes initial encoding through backbone layers. Subsequently, feature maps from the encoder layers are integrated into corresponding CFPFormer layers. Each decoder module partitions the input into patches, then applies Feature Re-encoding and Gaussian Attention, succeeded by a spatial resolution enhancement layer. Finally, the output of CFPFormer is fed into a segmentation head layer for final semantic segmentation.}
    \label{fig:model-architecture}
\end{figure*}

\subsection{Network Architecture}

The feature embeddings of the backbone serve as input to the Cross Feature Pyramid (CFP) block at the lowest resolution level of the pyramidal hierarchy, as shown in Fig. \ref{fig:model-architecture}. The output of this block is then upsampled and passed to the next CFP block at a higher resolution level in the pyramid. This process is repeated, progressively moving up the pyramid to higher resolution levels. 

At each level of the pyramid, the CFP block receives the upsampled features from the previous lower-resolution block. These upsampled features are combined with features from the same resolution level of the backbone network, which provides low-level spatial information and enhances long-range dependency to guide the attention mechanisms within the CFP block. 

As the decoding process ascends through the pyramidal hierarchy, the features are passed through Gaussian Attention, with the calculated decay mask based upon distance, and subsequently output is a set of high-resolution feature maps with strengthened mechanism to the top of the pyramid.

\subsection{Cross Feature Pyramid (CFP) Block}

\paragraph{\textbf{Pyramid Connection.}}
CFPFormer incorporates cross-layer interactions reminiscent of U-Net architectures. In line with other Vision Transformer (ViT) networks \cite{dosovitskiy_image_2021}, CFPFormer employs an attention mechanism with learnable Query ($\mathbf{Q}$), Key ($\mathbf{K}$), and Value ($\mathbf{V}$) matrices in each decoder layer. However, CFPFormer introduces a modification by augmenting the $\mathbf{K}$ and $\mathbf{V}$ matrices with additional features derived from the encoder layers. 

The  module first linearly projects the input embeddings $\mathbf{X' \in \mathbb{R}^{N \times D}}$ into query ($\mathbf{Q}$), key ($\mathbf{K}$), and value ($\mathbf{V}$) representations:

\begin{align}
Q = X'W^Q \\
K = X'W^K \oplus F_{enc} \\
V = X'W^V \oplus F_{enc}
\end{align}

where $\mathbf{W}^Q$, $\mathbf{W}^K$, and $\mathbf{W}^V$ are learnable linear layer matrices, and $F_{enc} \in \mathbb{R}^{H \times W}$ denotes the features extracted from the image encoder. It is important to note that patchmerging layers, commonly attached at the end of the bottleneck, reduce feature map dimensions while increasing channel dimensions. To address the resulting dimensional mismatch between encoder and decoder, we introduce Feature Re-encoding (FRE). FRE is designed to integrate diverse features within the attention layer, potentially facilitating more effective information flow between encoder and decoder.

\begin{figure}[htbp]
\centering
\subfloat[Pyramid Attention]{
    \includegraphics[width=0.15\textwidth]{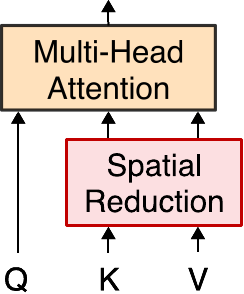} 
    \label{fig:pyramid-attention}
}
\hfill
\subfloat[Feature Re-encoding]{
    \includegraphics[width=0.25\textwidth]{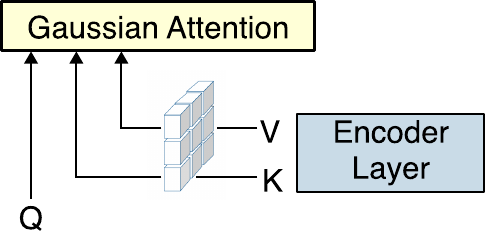} 
    \label{fig:feature-re-encoding}
}
\caption{Comparison between Pyramid Attention from PVT\cite{C:journals/corr/abs-2102-12122} and our proposed structure. Our Feature Re-encoding rearranges and interacts attention tensor K and V with image features from the backbone encoder.}
\label{fig:attentions}
\end{figure}

\begin{algorithm}
\caption{Gaussian Attention with Decomposed Axis Masks}
\label{alg:pseudocode}
\begin{algorithmic}[1]
\Require 
    \textbf{Input feature} $x$, 
    \textbf{Encoder layer} $\text{encoder\_layer}$, 
    \textbf{Row mask} $\text{mask\_h} \in \mathbb{R}^{H \times H}$, 
    \textbf{Column mask} $\text{mask\_w} \in \mathbb{R}^{W \times W}$
\Ensure 
    \textbf{Output feature} $\text{output} \in \mathbb{R}^{C}$

\Procedure{Attention}{$x$, $\text{encoder\_layer}$, $\text{mask\_h}$, $\text{mask\_w}$}
    \State $q, k, v \gets \text{SpatialReduction}(x, \text{encoder\_layer})$ 
    \State $\text{lepe} \gets \text{PositionalEncoding}(v)$ 
    \State $qr\_w, kr\_w, v \gets \text{PermuteForRow}(q, k, v)$ \Comment{Permute for row-wise attention}

    \State $A_r \gets \text{matmul}(\text{log\_softmax}(qr\_w \cdot kr\_w^T + \text{mask\_w}, -1), v)$ \Comment{Row-wise attention}
    \State $A_r \gets A_r + \text{mask\_w}$ \Comment{Apply row mask}
    \State $v\_w \gets A_r @ v$ \Comment{Update values with row attention}

    \State $qr\_h, kr\_h, v \gets \text{PermuteForCol}(q, k, v)$ \Comment{Permute for column-wise attention}

    \State $A_c \gets \text{matmul}(\text{log\_softmax}(qr\_h \cdot kr\_h^T + \text{mask\_h}, -1), v\_w)$ \Comment{Column-wise attention}
    \State $A_c \gets A_c + \text{mask\_h}$ \Comment{Apply column mask}
    \State $\text{output} \gets A_c @ v + \text{lepe}$ \Comment{Combine with positional encoding}

    \State \textbf{return} $\text{output}$ \Comment{Return final output}
\EndProcedure
\end{algorithmic}
\end{algorithm}

\paragraph{\textbf{Feature Re-encoding.}}

A key component of the CFP block is the Feature Re-encoding mechanism, which aims to enhance the model's ability to capture fine-grained details and small structures by leveraging information from lower-resolution feature maps.

To incorporate cross-layer feature information, the key (K) and value (V) tensors in the Feature Re-encoding (FRE) module are combined with encoder features $F_\text{enc} $ from a lower layer of the network, as depicted in follows:

\begin{equation}
FRE(K,F_{enc}) = 
    FRE(V,F_{enc}) = V \oplus Patchembed(F_{enc})
\label{eq8}
\end{equation}

where $ F_\text{enc} \in \mathbb{R}^{B \times H_\text{enc} \times W_\text{enc} \times C_\text{enc}}$, Patchembed are layers decomposing features into image embeddings. Here we accept the features in down-sampling layers, with image size $H_{enc}$ with $W_{enc}$, varying into $\frac{H_{enc}}{P}$ with $\frac{W_{enc}}{P}$. Unlike those cascaded decoder, for example: TransUnet, PVT-CASCADE, which directly concatenate up-sampling convolutional layers with encoder features, the cross-feature combination allows the attention mechanism to interact with low-level spatial information from the encoder in a more effective way. A similar work can be traced to Pyramid Vision Transformers\cite{C:journals/corr/abs-2102-12122}, from which suggest using Spatial Reduction to fit channel dimensions of K and V tensors by linear projection. 

However, our method connects pyramid layers in between encoder and decoder layers, which enables the model to better capture fine-grained details by taking advantages of Gaussian Attention and structures present in the input data, as depicted in Fig. \ref{fig:attentions}. Further experiments on the impact of FRE and connections of pyramid layers are demonstrated in \nameref{chap:ablation}.

\paragraph{\textbf{Gaussian Attention.}}

\begin{figure}[h]
    \centering
    \includegraphics[width=0.8\linewidth]{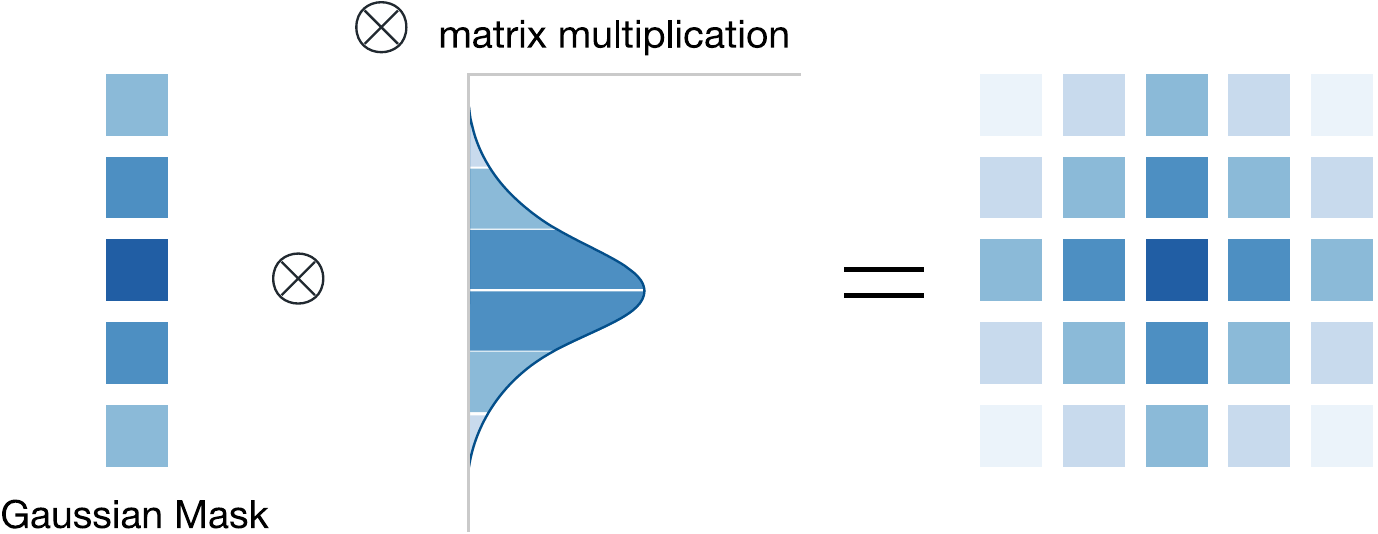}
    \caption{Gaussian Attention Decay. GA module first initialize decay function in the beginning of the model, and thereby calculate distance between each point to generate context masking. Eventually masks are applied after MHSA operation then conveyed image patches into next layer.}
    \label{fig:gaussian-attention}
\end{figure}

As shown in pseudocode algorithm \ref{alg:pseudocode}, the CFP block employs an attention mechanism, termed Gaussian Attention(GA) module, which situate the feature computations along rows and columns in pixels. In order to mitigate the burdening computational cost triggered by most of Vision Transformers\cite{chen2024comprehensive,  heidari2024enhancing} and redundancies of Multi-head Self Attention(MHSA), GA module retains context feature in a certain decay pattern, including those preceding and following. For each token, the output of attention demonstrates in Eq. \ref{eq:GA}.

\begin{align}
\label{eq:GA}
\mathrm{GaussianAttention}(X) =(QK^{\mathsf{T}}\odot D)V \\
D_{nm} =\gamma^{|n-m|} 
\end{align}

\noindent where D denotes the decay generated from GA module and it multiplies with $Q$ and $K$ during each MHSA operation. Fig. \ref{fig:gaussian-attention} illustrates the decay pattern of Gaussian Attention, showing how the attention weight decreases as the distance from the central point increases.

\vspace{-2mm}

\paragraph{\textbf{Axial Decomposed Calculation for Attention Decay.}}
 Inspired by masking decays Retentive Vision Transformer\cite{fan2023rmtretentivenetworksmeet}, we adopt Axial Decomposed Calculation upon Gaussian decay, from which our module efficiently dissociates all-pixel wised attention into row-wised attention and column-wised attention. Through this bi-directional operation, the decay pattern becomes more suitable for image downstream tasks.

GA module first learns inter-patch multi-head correlations and are then reshaped and used for row-wise and column-wise attention computations:

\begin{align}
A_r &= \text{softmax}\left(\frac{\text{reshape}(Q) \text{reshape}(K)^T}{\sqrt{d_k}}\right) \odot M \\
A_c &= \text{softmax}\left(\frac{\text{reshape}(Q) \text{reshape}(K)^T}{\sqrt{d_k}}\right) \odot M  
\end{align}

where $A_r$ and $A_c$ denotes row-wise attention and column attention respectively, while $\text{reshape}(\mathbf{Q})$, $\text{reshape}(\mathbf{K})$, and $\text{reshape}(\mathbf{V})$ are reshaped matrices suitable for attention computations, and $d_k$ is the dimensionality of the key vectors.

Unlike attention mechanisms that assign equal importance to all positions within the receptive field, Gaussian attention decays the attention weights based on a Gaussian curve, efficiently prioritizing information from relevant image patches while filtering out noise and irrelevant details.

The Gaussian attention mechanism is implemented by generating a 2D decay mask $\mathbf{M} \in \mathbb{R}^{H' \times W'}$ based on the Euclidean distance between spatial positions:

\begin{equation}
\mathbf{M}[i, j] = \exp\left(-\frac{i^2 + j^2}{2\sigma^2}\right)
\label{eq6}
\end{equation}

\begin{equation}
D = M^ {\sqrt{(x_2 - x_1)^2 + (y_2 - y_1)^2}}
\label{eq7}
\end{equation}
where $\sigma$ is a learnable parameter that controls the rate of decay.

This decay mask is then applied to the attention scores, effectively modulating the attention weights with a Gaussian decay, as depicted in equation \ref{eq7}.

\section{Experiments}
\subsection{Datasets}

To evaluate the effectiveness of our proposed CFPFormer method, we performed experiments on Medical Image Segmentation.

\textbf{Medical Image Segmentation Datasets.} we employed two challenging datasets: the MRI Automatic Cardiac Diagnosis Challenge (ACDC) \cite{bernard_deep_2018} and the Synapse Multi-organ Segmentation Challenge \cite{infosagebaseorg_synapse_nodate}. The ACDC dataset comprises 100 MRI scans, with ground truth annotations for the left ventricle (LV), right ventricle (RV), and myocardium (MYO). We followed a standard train-validation-test split of 70-10-20. The Synapse dataset, on the other hand, contains CT scans from 30 patients, and our experimental setup and pre-processing closely followed the methodology described in TransUNet\cite{chen_transunet_2021}. We also followed their data argumentation which and during the data augmentation process, we incorporated random rotations of 0, 90, 180, or 270 degrees, as well as horizontal or vertical flips, each with a 50\% probability. Additionally, we resized the images using cubic interpolation to attain a specific image size.

In the case of Medical Image Segmentation, we utilized the widely-used Dice Similarity Coefficient (DSC) and Hausdorff Distance (HD) metrics to assess the model's performance. The DSC measures the overlap between the predicted segmentation masks ($P$) and the ground truth masks ($G$), and is defined as:

\begin{equation}
\text{DSC}(P, G) = \frac{2 \times |P \cap G|}{|P| + |G|}
\end{equation}

where $|\cdot|$ denotes the cardinality of a set. A DSC value of 1 indicates perfect overlap between the predictions and ground truth.

The Hausdorff Distance (HD) quantifies the maximum distance between the predicted and ground truth boundaries, and is calculated as:

\begin{equation}
\text{HD}(P, G) = \max \left\{ \sup_{p \in P} \inf_{g \in G} d(p, g), \sup_{g \in G} \inf_{p \in P} d(g, p) \right\}
\end{equation}

where $d(p, g)$ represents the Euclidean distance between points $p$ and $g$. A lower HD value indicates better alignment between the predicted and ground truth boundaries.

\subsection{Model Settings}

Our proposed CFPFormer model demonstrates flexibility in terms of block ratio and number. We developed two primary variants: CFPFormer-Tiny (CFPFormer-T) and CFPFormer-Small (CFPFormer-S). Table~\ref{tab:architecture} delineates the specific parameter settings for each variant.

\begin{table}[ht]
\centering
\renewcommand{\arraystretch}{1.01}
\setlength{\tabcolsep}{3pt}  
\resizebox{\columnwidth}{!}{  
\begin{tabular}{l|c|c|c|c}
\toprule
Model & $C_1$, $C_2$, $C_3$, $C_4$ & MLP Ratio & $A_1$, $A_2$, $A_3$, $A_4$ & Drop Rate \\
\midrule
CFPFormer-T & 1, 1, 3, 1 & 3 & 2, 4, 8, 16 & 0.15 \\
CFPFormer-S & 2, 2, 6, 2 & 3 & 4, 4, 8, 16 & 0.20 \\
\bottomrule
\end{tabular}
}
\caption{Model Parameter Settings of CFPFormer variants.}
\label{tab:architecture}
\end{table}

In the default configuration, CFPFormer-T, the bottleneck blocks are set to 1, 1, 3, 1, representing the number of blocks in each stage ($C_1$, $C_2$, $C_3$, $C_4$). To mitigate overfitting, we implemented a drop-path rate of 0.15. The attention heads in each stage ($A_1$, $A_2$, $A_3$, $A_4$) are set to 2, 4, 8, and 16, respectively.

CFPFormer-S, a comparatively larger model, incorporates several enhancements over its predecessor. Specifically, we increased the depth parameters in each bottleneck layer (2, 2, 6, 2) and doubled the number of attention heads in the initial self-attention layer (4, 4, 8, 16). To account for the increased model complexity, we adjusted the drop-path rate to 0.20.

Both variants maintain a consistent MLP ratio of 3 across all stages, balancing model capacity and computational efficiency.
\subsection{Training Process}

For both the ACDC~\cite{bernard_deep_2018} and Synapse~\cite{infosagebaseorg_synapse_nodate} datasets, we employed a consistent image size of $256\times256$. The training process was initiated with a learning rate of $1e-4$, incorporating a decay rate of $1e-4$. To optimize our models during training, we utilized the Adam optimizer~\cite{kingma2014adam}.

During the training phase, we optimized our models using the Adaptive t-vMF Dice Loss, as proposed by Kato et al.~\cite{kato2022adaptive}. This loss function has demonstrated efficacy in medical image segmentation tasks, particularly in handling class imbalance and complex anatomical structures.

While larger parameter sets in transformer model series can potentially lead to improved performance, we observed that they also increased the risk of underfitting and training instability. These issues could potentially decrease overall performance. Consequently, we focused our experiments on the two variants described above: CFPFormer-T and CFPFormer-S.

The training process was carefully monitored to ensure stability and convergence, with particular attention paid to the larger CFPFormer-S variant. Our experimental setup aims to evaluate the performance of both models across various medical image segmentation tasks, leveraging the ACDC and Synapse datasets. This approach allows us to assess the model's versatility and efficiency across different anatomical structures and imaging modalities.

\subsection{Results}

For the Medical Image Segmentation tasks, we report the DSC and HD metrics for the ACDC and Synapse datasets in Tables \ref{tab:acdc_results} and \ref{tab:synapse_results} respectively.For evaluation usage, our decoder is assembled with U-net as the encoder, representative as a baseline model of Medical Segmentation, coupled with backbones VGG-16 and Resnet-50. 

\begin{table}[htbp]
    \centering
    \small  
    \renewcommand{\arraystretch}{1.1}  
    \setlength{\tabcolsep}{3pt}         
    \resizebox{\columnwidth}{!}{         
    \begin{tabular}{llcccc}
    \toprule
    Type & Method & DSC (LV) & DSC (RV) & DSC (MYO) & Avg. \\
    \midrule
    
    \multirow{5}{*}{CNN} 
    & R50 U-Net \cite{ronneberger_u-net_2015} & 87.10 & 80.63 & 94.92 & 87.55 \\
    & R50 ViT & 86.07 & 81.88 & 94.75 & 87.57 \\
    & EMCAD \cite{rahman2024emcad} & 89.37 & 88.99 & 95.65 & 91.34 \\
    & \textbf{VGG-16 CFPFormer-T (Ours)} & 87.91 & 88.46 & 95.10 & 90.49 \\
    & \textbf{R50 CFPFormer-T(Ours)} & 87.92 & \textbf{89.54} & \textbf{95.93} & 91.10 \\
    & \textbf{R50 CFPFormer-S(Ours)} & 89.03 & \textbf{89.37} & \textbf{95.68} & \textbf{91.36} \\
    \midrule
    \multirow{7}{*}{Transformer}
    & ViT \cite{dosovitskiy_image_2021} & 81.46 & 70.71 & 92.18 & 81.45 \\
    
    & ViT+CUP \cite{dosovitskiy_image_2021} & 81.46 & 70.71 & 92.18 & 81.45 \\
    & TransUNet \cite{chen_transunet_2021} & 89.71 & 88.86 & 84.53 & 89.71 \\
    & SwinUNet \cite{cao_swin-unet_2023} & 88.55 & 85.62 & 95.83 & 90.00 \\
    & PVT-CASCADE \cite{rahman2023medical} & 88.90 & 89.97 & 95.50 & 91.46 \\
    & $FCT_{224}$ \cite{tragakis2023fully} & 92.02 & 90.61 & 95.89 & 92.84 \\
    
    & SegFormer \cite{perera2024segformer3d} & 88.50 & 88.86 & 95.53 & 90.96\\
    & \textbf{PVT CFPFormer-S(Ours)} & 89.95 & \textbf{90.11} & \textbf{96.02} & \textbf{92.02} \\
    \bottomrule
    \end{tabular}
    }
    \caption{ACDC dataset segmentation results}
    \label{tab:acdc_results}
\end{table}

Our CFPFormer architecture outperformed other relevant methods, especially from its baseline U-net, demonstrating its effectiveness in capturing intricate anatomical structures and delineating precise segmentation boundaries. Compared with those models with Resnet-50 Backbone, yet our VGG-16 CFPFormer, with less parameters in backbone, evidently exceed in both RV and MYO categories. Our R50 CFPFormer, making use of strong context extraction and larger parameter size, reaches a better DSC in MYO Category.

\begin{table}[H]
\centering

\begin{tabular}{lcc}
\toprule
Method & DSC & Liver \\
\midrule
MT-Unet \cite{wu2023mtu} & 78.59 & 93.06  \\
R50 U-Net \cite{ronneberger_u-net_2015} & 74.68 & 93.35 \\
R50 Att-Net \cite{ronneberger_u-net_2015} & 75.57 & 93.56 \\
TransUNet \cite{chen_transunet_2021} & 77.48 & 94.08 \\
SwinUNet \cite{cao_swin-unet_2023} &\textbf{ 79.13} & 94.29  \\
\textbf{VGG-16 CFPFormer-T (Ours)} & 78.45 & \textbf{94.85} \\
\bottomrule
\end{tabular}
\caption{Synapse dataset segmentation results}
\label{tab:synapse_results}
\end{table}

\begin{figure*}[h!]
\centering
\scalebox{0.8}{%
\begin{subfigure}{0.3\textwidth}
\includegraphics[width=\linewidth]{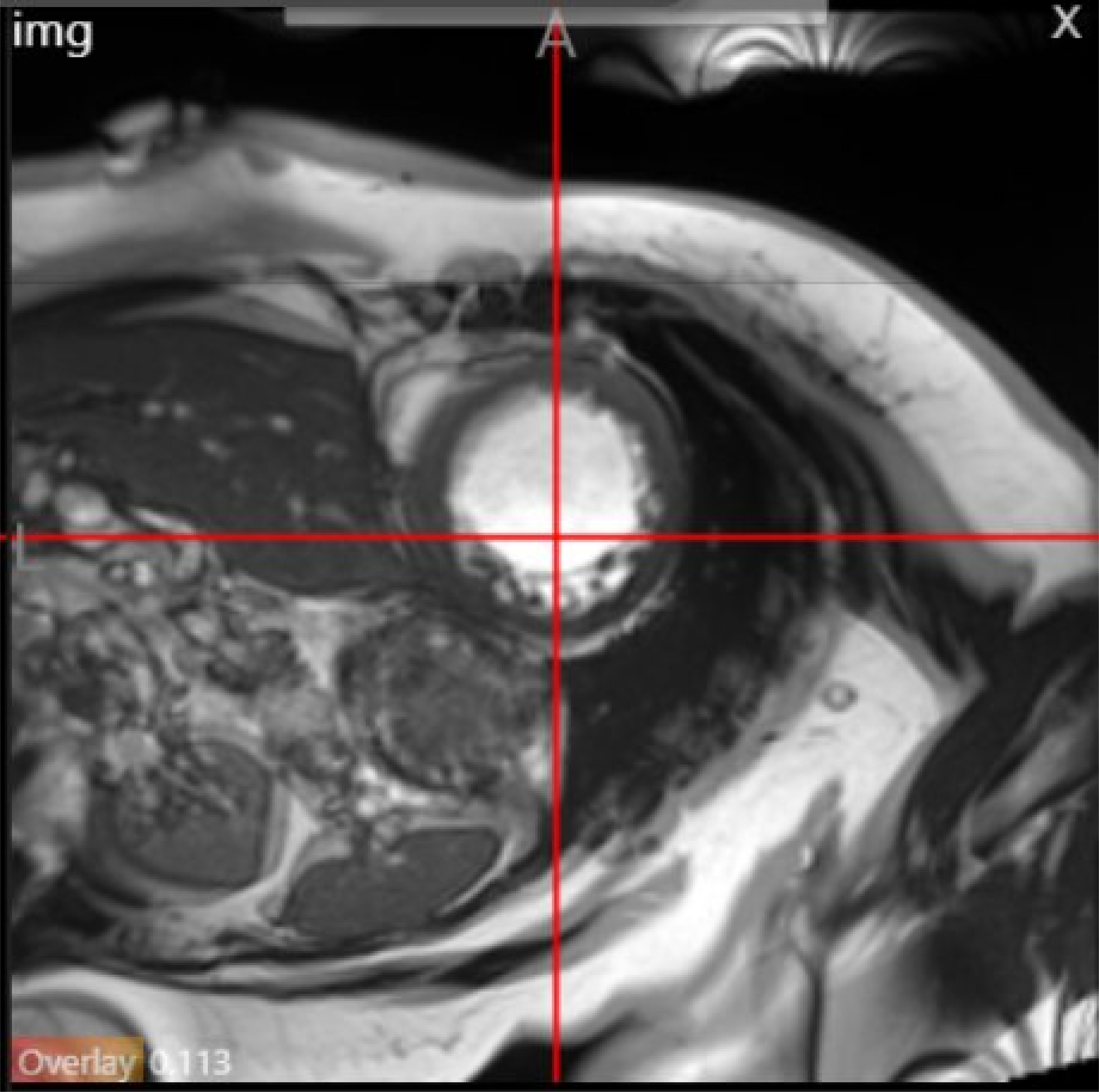}
\caption{Input Image}
\label{fig:input-image}
\end{subfigure}%
\hfill
\begin{subfigure}{0.3\textwidth}
\includegraphics[width=\linewidth]{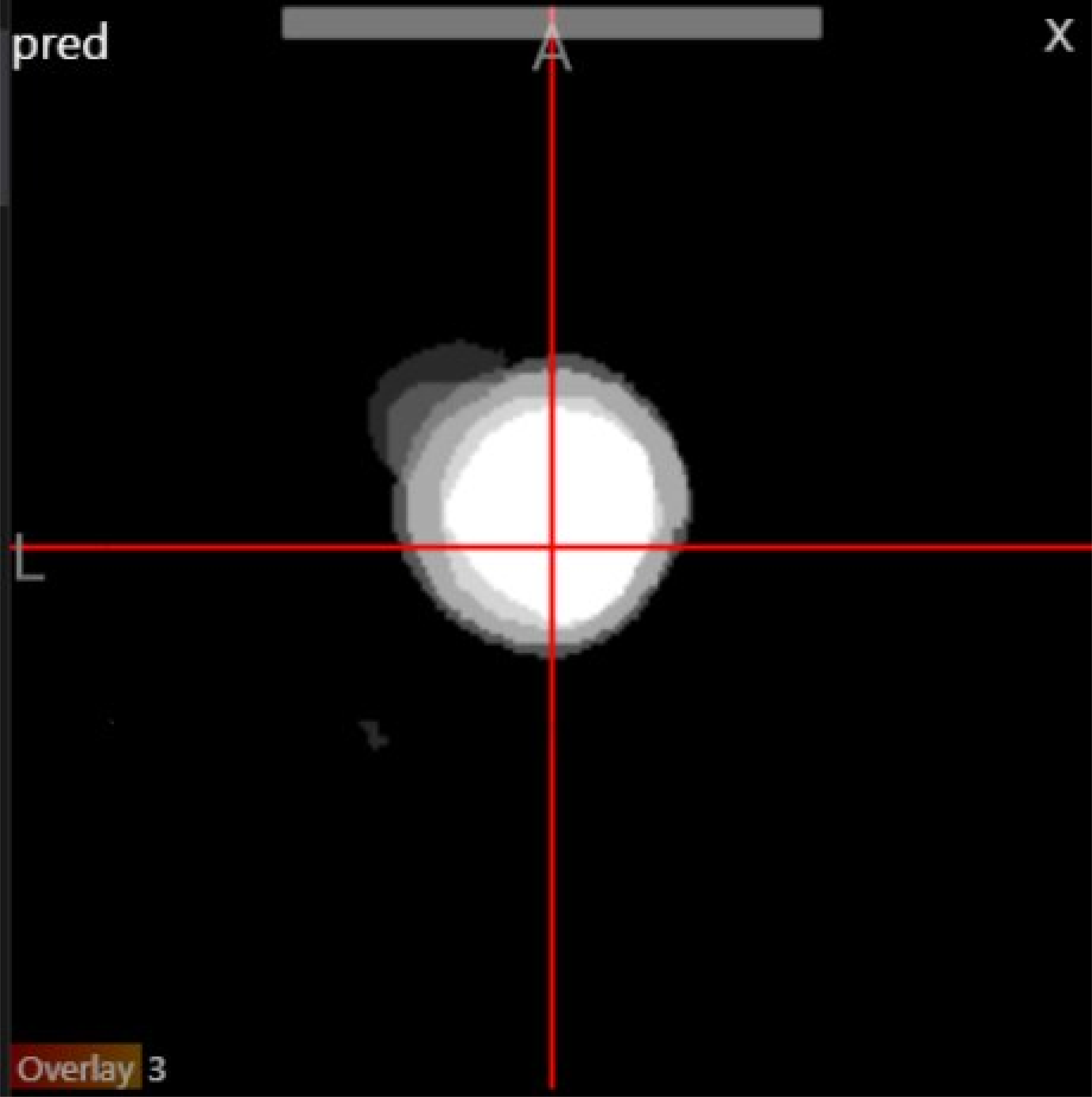}
\caption{Prediction}
\label{fig:prediction}
\end{subfigure}%
\hfill
\begin{subfigure}{0.3\textwidth}
\includegraphics[width=\linewidth]{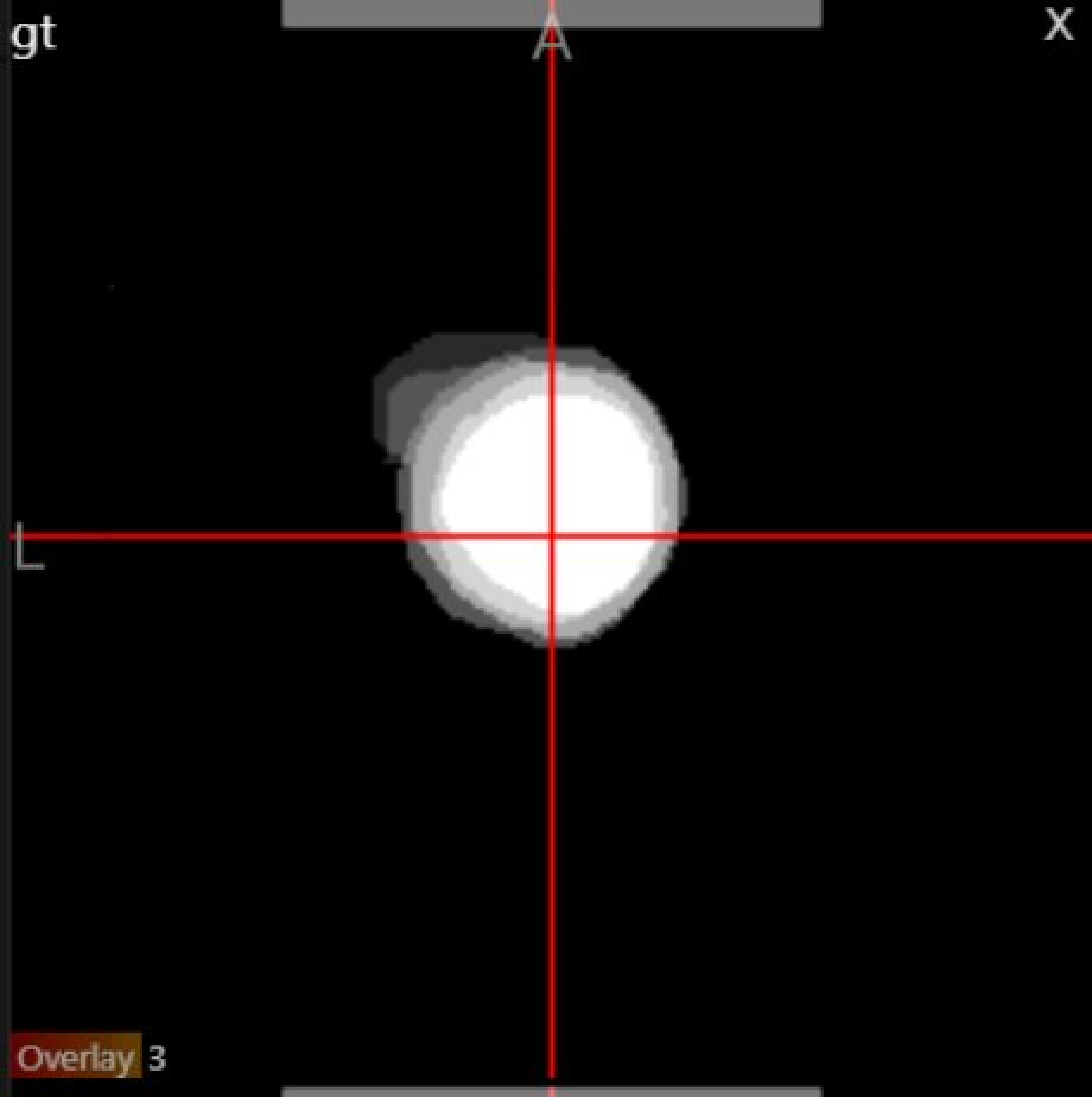}
\caption{Ground Truth}
\label{fig:ground-truth}
\end{subfigure}
}
\caption{ACDC segmentation example 1: (a) Original MRI slice, (b) Prediction result of CFPFormer, and (c) Ground truth segmentation masks.}
\label{fig:acdc_vis}
\end{figure*}

\begin{figure*}[h!]
\centering
\scalebox{0.8}{%
\begin{subfigure}{0.3\textwidth}
\includegraphics[width=\linewidth]{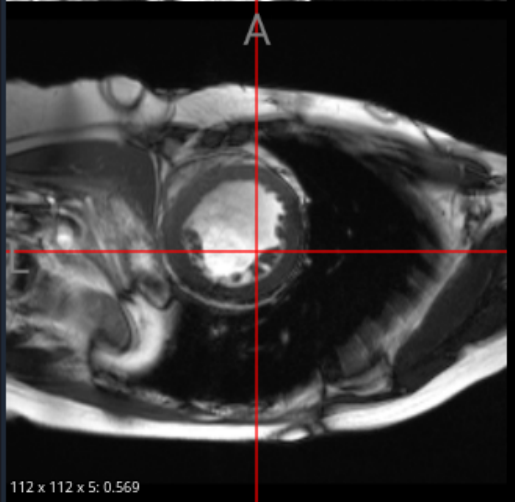}
\caption{Input Image}
\label{fig:input-image}
\end{subfigure}%
\hfill
\begin{subfigure}{0.3\textwidth}
\includegraphics[width=\linewidth]{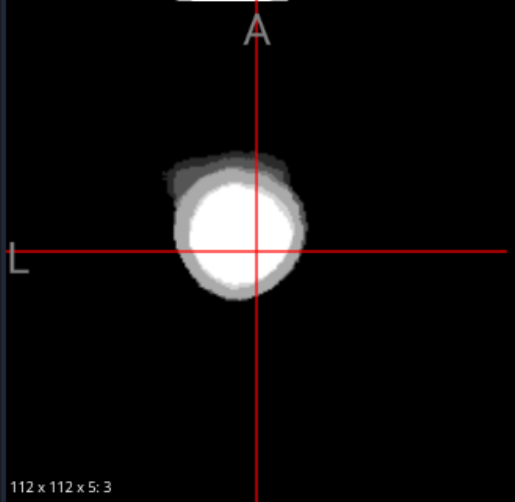}
\caption{Prediction}
\label{fig:prediction}
\end{subfigure}%
\hfill
\begin{subfigure}{0.3\textwidth}
\includegraphics[width=\linewidth]{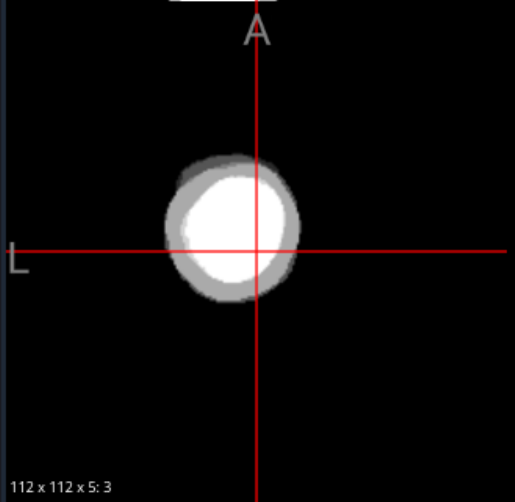}
\caption{Ground Truth}
\label{fig:ground-truth}
\end{subfigure}
}
\caption{ACDC segmentation example 2: (a) Original MRI slice, (b) Prediction result of CFPFormer, and (c) Ground truth segmentation masks.}
\label{fig:acdc2}
\end{figure*}

To provide an evident evaluation, Fig. \ref{fig:acdc_vis} and Fig. \ref{fig:acdc2} illustrates a sample MRI slice from the ACDC dataset, along with the corresponding prediction and ground truth segmentation masks generated by our CFPFormer model. The given figures demonstrate the model's ability to accurately segment intricate anatomical structures, such as the left and right ventricles, and the myocardium. 

\subsection{Analysis}
\textbf{Increased Accuracy in the Downstream Tasks.} Our method proves a better performance on average compared with baseline models. Since our main idea is to improve the model during the decoding part, we integrate our CFPFormer into the upsampling layers of U-Net, which we consider areas that need improvement. The results shown in Table \ref{tab:acdc_results} demonstrate the superior performance achieved by adopting our methods.

\textbf{Boosting with Transformer Encoder.} We combine Pyramid Vison Transformer\cite{wang2021pyramid} and examine on ACDC datasets in Table \ref{tab:acdc_results}. PVT primarily acts as image backbone nowadays and performed a higher accruacy than Resnet50. To stack with a larger encoder, especially transformer-based networks, we easily plug CFPFormers and pass encoder features through training functions. In order to reduce the complexity during the model construction, we use a set of parameters to receive tensors from each encoder layer. Our CFPFormer decoder consists of 4 blocks transformer decoders, unlike hybrid architectures like TransUNet that require retraining the entire encoder-decoder stack, CFPFormer acts as a plug-and-play decoder. As a result, PVT CFPFormer-T outperforms PVT-CASCADE\cite{rahman2023medical} by 0.57 points in DSC, which proves the better decomposition and rearrangement than CASCADE Decoder\cite{rahman2023medical}.

\subsection{Ablation Studies}
\label{chap:ablation}

\textbf{Upsampling Layers.} A slight difference can be spotted from choosing different upsampling layers. Here we experiment on Transpose Convolutional Layer and Bilinear Interpolation, as shown in \ref{tab:upsampling}. The results indicates superior upsampling capability, while taking up for less parameters at the same time. 

\begin{table}
\centering
\scalebox{0.92}{
\begin{tabular}{c|c|c|c}
\toprule
Upsampling Methods & Backbone & Params(M) & DSC Avg.  \\
\midrule

\textbf{CFPFormer-T w/ TransposeConv} & VGG-16 &  222.5 & \textbf{89.53} \\
\textbf{CFPFormer-T w/ Blinear} & VGG-16 & 221.7 & \textbf{90.49} \\
\bottomrule
\end{tabular}
}
\caption{Ablation studies on upsampling layers}
\label{tab:upsampling}
\end{table}

\textbf{Comparison with related works.}
To further analyze the impact of various components in our CFPFormer architecture, we strictly compare with previous related works, by aligning our model into existing models as a decoder.

\begin{table}[h]
\centering

\scalebox{1.00}{
\begin{tabular}{lc |c}
\toprule
Method & Params(M)  & DSC Avg.\\
\midrule
VGG-16 CFPFormer-T  & 221.7 & 91.10 \\
GA $\rightarrow$ MHSA &  196.6  & 89.51\\ 
w/o Pyramid Connection  & 221.6 & 90.16 \\
$Log_2$Softmax $\rightarrow$ Softmax & 221.7 & 90.49 \\
w/o FRE  & 221.6 & Nan \\

\bottomrule
\end{tabular}
}
\caption{Ablation studies conducted on ACDC dataset}
\label{tab:ablation_studies}
\end{table}

\textbf{Gaussian Attention \& Softmax.}
The row "CFPFormer w/o GA" refers to the CFPFormer model without the Gaussian Attention component. Instead, we replace Gaussian Attention with default settings of Multi-head Self Attention (MHSA)\cite{vaswani_attention_2023}. The result shows that GA module rises the model's performance in ACDC task. In terms of attention score, we migrate softmax to log2softmax to improve the calculation.

\textbf{Feature Re-encoding with K \& V.}
The row "CFPFormer w/o FRE" refers to the CFPFormer model without the Feature Re-encoding with Key (K) and Value (V) component and Pyramid Connection between encoder layers. Without the aid of FRE module, the model is likely to fail when Pyramid Connection stage.

\textbf{Pyramid Connection}
The involvement of Pyramid Connection gains more insights from image original features. Our conducted ablation experiments indicate that this extra operation facilitates loss caused by multiple blocks of the decoder.



\section{Conclusion}
Our work mainly contributes a novel decoder that associates the features across with Encoder layers, with U-shaped Pyramid Re-encoding connections between modules, which is conducive to undermine the deterioration of feature lost caused by those long distant models. Our Gaussian Attention mechanism successfully speeds up the computation while scaling up in the model, and effectively utilized masked decays of Gaussian distribution to elevate the performance taken from Attention.

\section{Acknowledgment}
This research was supported in part through computational resources provided by the Data-Intensive Computing Centre, Universiti Malaya.

\bibliography{relevantwork2}
\end{document}